\documentclass[preprint]{elsarticle}

\usepackage{lineno}
\usepackage{framed,multirow}

\usepackage{amssymb}
\usepackage{latexsym}
\usepackage{times}
\usepackage{xcolor}
\usepackage{soul}
\usepackage[utf8]{inputenc}
\usepackage[small]{caption}
\usepackage{mathtools}
\usepackage{amsfonts}
\usepackage{amsmath,graphicx}
\usepackage{epstopdf}
\usepackage{makecell}
\usepackage{subfig}
\usepackage{diagbox}
\usepackage{ulem}
\newcommand{\PreserveBackslash}[1]{\let\temp=\\#1\let\\=\temp}
\newcolumntype{C}[1]{>{\PreserveBackslash\centering}p{#1}}
\newcolumntype{R}[1]{>{\PreserveBackslash\raggedleft}p{#1}}
\newcolumntype{L}[1]{>{\PreserveBackslash\raggedright}p{#1}}
\usepackage{url}

\journal{Journal of \LaTeX\ Templates}

\begin{document}

\begin{frontmatter}

\title{Weighted Bilinear Coding over Salient Body Parts for Person Re-identification}


\author[mymainaddress]{Zhigang Chang}
\author[mymainaddress,myfifthaddress]{Zhou Qin\tnoteref{mytitlenote}}

\author[mysecondaryaddress]{Heng Fan}
\author[myfourthaddress]{Hang Su}
\author[mymainaddress]{Hua Yang}

\author[mymainaddress]{Shibao Zheng\corref{mycorrespondingauthor}}
\cortext[mycorrespondingauthor]{Corresponding author}
\ead{sbzh@sjtu.edu.cn}
\author[mysecondaryaddress,mythirdaddress]{Haibin Ling}

\address[mymainaddress]{Institute of Image Processing and Network Engineering, Shanghai Jiao Tong University,  Shanghai 200240, China}
\address[mysecondaryaddress]{Department of Computer \& Information Sciences, Temple University, Philadelphia 19122, USA}
\address[mythirdaddress]{Computer Science and Engineering, South China University of Technology, Guangzhou 510006, China}
\address[myfourthaddress]{Department of Computer Science and Technology, Tsinghua University, Beijing 100084, China}
\address[myfifthaddress]{Artificial Intelligence Center-City Brain, Alibaba Cloud, Hangzhou 311100, China}

\begin{abstract}
Deep convolutional neural networks (CNNs) have demonstrated dominant performance in person re-identification (Re-ID). Existing CNN based methods utilize global average pooling (GAP) to aggregate intermediate convolutional features for Re-ID. However, this strategy only considers the first-order statistics of local features and treats local features at different locations equally important, leading to sub-optimal feature representation. To deal with these issues, we propose a novel \emph{weighted bilinear coding} (WBC) framework for local feature aggregation in CNN networks to pursue more representative and discriminative feature representations, which can adapt to other state-of-the-art methods and improve their performance. In specific, bilinear coding is used to encode the channel-wise feature correlations to capture richer feature interactions. Meanwhile, a weighting scheme is applied on the bilinear coding to adaptively adjust the weights of local features at different locations based on their importance in recognition, further improving the discriminability of feature aggregation. To handle the spatial misalignment issue, we use a salient part net (spatial attention module) to derive salient body parts, and apply the WBC model on each part. The final representation, formed by concatenating the WBC encoded features of each part, is both discriminative and resistant to spatial misalignment. Experiments on three benchmarks including Market-1501, DukeMTMC-reID and CUHK03 evidence the favorable performance of our method against other outstanding methods.

\end{abstract}

\begin{keyword}
Person Re-identification \sep Bilinear coding
\end{keyword}

\end{frontmatter}

\section{Introduction}

Person re-identification (Re-ID)~\cite{survey/ZhengYH16, hermans2017defense, sun2018beyond, tao2016person, tao2013person, wang2018learning} aims at associating a probe image with images of the same identity in the gallery set (usually across different non-overlapping camera views). It is attracting increasing attentions due to its importance for various applications including video surveillance, human-machine interaction, robotics, etc. Despite years of efforts, accurate Re-ID remains largely unsolved because of great challenges posed by illumination changes, pose variations or viewpoint changes, and other factors like background clutters and occlusions. Various techniques have been proposed to improve the recognition performance against the above-mentioned challenges.

Motivated by the success in image classification~\cite{googlenet/SzegedyLJSRAEVR15,resnet/HeZRS16,vggnet/SimonyanZ14a}, semantic segmentation~\cite{long2015fully} and tracking~\cite{fan2017parallel,fan2017sanet}, convolutional neural networks (CNNs)~\cite{lecun1989backpropagation} have been widely utilized for Re-ID because of its power in learning discriminative and representative features. Being an end-to-end architecture, CNNs directly take as input the raw images, and hierarchically aggregate local features into a final vectorized representation for further processing. In such Re-ID solutions, one crucial problem is how to aggregate the intermediate convolutional features to build more discriminative appearance representation for better recognition performance. For the sake of efficiency and simplicity, most CNN based approaches use global average pooling (GAP) to aggregate the convolutional features to represent human appearance~\cite{deep_aligned/ZhaoLZW17,pose_driven/SuLZX0T17}. However, discarding the information of feature correlations as well as the various feature importance across different locations, GAP leads to suboptimal aggregated appearance representation.

To deal with this issue, in this paper, we propose a novel \emph{weighted bilinear coding} (WBC) framework for discriminative feature aggregation in CNNs, which is able to model richer higher-order feature interactions as well as the various feature impacts for Re-ID. The superiority of our WBC framework comes from two aspects: Firstly, the bilinear coding takes into consideration the channel-wise correlations of each local feature. In comparison to global average pooling, bilinear coding captures richer feature information. More importantly, considering that the features at different locations have different impacts on the recognition performance, we further introduce a weighting scheme into bilinear coding, which adaptively weighs different features according to their relative importance in recognition. 

The proposed WBC framework is flexible and can be embedded into arbitrary networks as a feature aggregation part. The WBC framework provides a new template for higher-order feature aggregation and the popular Re-ID models such as Refined Part Pooling (PCB-RPP)~\cite{sun2018beyond} assembled with WBC get a significant improvement in performance compared to the original models.

\begin{figure*}[!t]
	\centering
	\includegraphics[width=\linewidth]{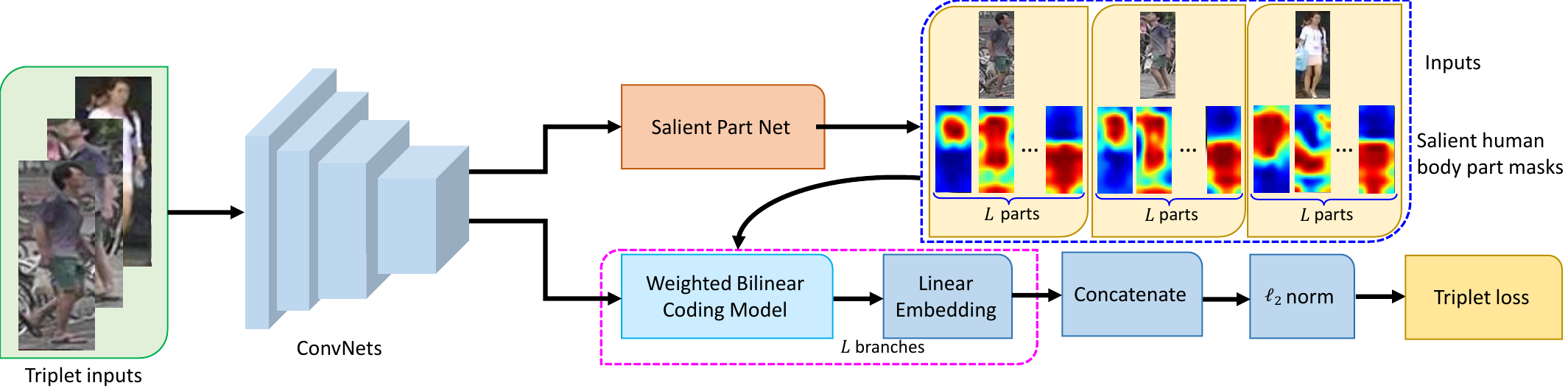}
	\captionsetup{font={footnotesize}}
	\vspace{-1ex}
	\caption{Illustration of the proposed person Re-ID framework. Based on the feature maps extracted from the backbone ConvNets, we first adopt a salient part net to obtain salient human body parts, then the proposed WBC model is applied on each part for discriminative feature aggregation. The final representation of each person is formed by concatenating the features of all parts, followed by $\ell_2$ normalization. Triplet loss calculated on the final representations is adopted for parameter learning of the Re-ID network.}\label{fig:proposed_model}
	\vspace{-2ex}
\end{figure*}

To deal with the problem of spatial misalignment in Re-ID, we integrate the proposed WBC model with a salient part net to pursue part-aligned discriminative representation for Re-ID. In specific, the salient part net is used to derive several salient human body parts, then we apply the proposed WBC on each part to obtain corresponding discriminative feature representation. The final representation for each human image, formed by concatenating the features of each part, bears the properties of both discriminability and resistance to spatial misalignment. So the representations over the parts are learned end to end and the similarities between the corresponding parts are aggregated. Therefore, each branch of the salient part net can learn attention mask of a local area of original feature map due to feature concatenation and triplet loss learning. The proposed Re-ID framework with salient part net and WBC, is illustrated in Figure \ref{fig:proposed_model}.

In summary, we make the following contributions:
\begin{itemize}
	\item We propose a novel framework for representative and discriminative feature aggregation considering channel-wise correlations of aligned local features, which can be flexibly plugged into existing deep architectures.
	
	\item To alleviate the spatial misalignment problem, we integrate the WBC model with a salient part network to pursue part-aligned higher-order interacted features in an end-to-end trainable network for Re-ID.
	
	\item Extensive experiments on three large-scale datasets including Market-1501~\cite{market_dataset/ZhengSTWWT15}, DukeMTMC-reID~\cite{duke_dataset/RistaniSZCT16} and CUHK03~\cite{cuhk03_dataset/LiZXW14} demonstrate the favorable performance of our framework against state-of-the-art approaches. Some experiments demonstrate that the popular Re-ID models with WBC outperforms the original models, which has proven its flexibility and generalization.
\end{itemize}

The rest of this paper is organized as follows. Section 2 reviews the relevant works of this paper. Section 3 illustrates the details of the proposed Re-ID method. Section 4 demonstrates the experimental results, followed by conclusion in Section 5.

\section{Related Work}
Being extensively studied, numerous approaches have been proposed for Re-ID in recent years~\cite{survey/ZhengYH16}. Sorted from the perceived scale of the extracted features, the Re-ID methods can be generally summarized into two categories: global-based models which take the whole human body into consideration during feature design or metric learning and part-based ones that extract features from local body parts and then aggregate these local features for final ranking. Earlier works mainly focus on global models for Re-ID~\cite{prdc/ZhengGX11,kissme/KostingerHWRB12,kernel/XiongGCS14}. These methods, nevertheless, degrade in presence of spatial misalignment caused by large variations in view angles and human poses.

Due to the inaccuracy of the foreground boxes of the pedestrian, such as scale problems, occlusion, environmental noise, etc., the aggregated parts features provide higher discriminative ability than global features. 

\subsection{Part-based Re-ID}
To alleviate the problem of spatial misalignment, many part-based algorithms have been proposed. Given that the human body is usually centered in a manually cropped bounding
boxes, some researchers argue that the body parts are vertically roughly aligned. Therefore a possible solution is to decompose human body into uniform stripes, and pool features extracted from these stripes into a robust representation. In~\cite{scsp/ChenYCZ16}, the authors propose to learn a sub-similarity function for each stripe, and then fuse all sub-similarity scores for final recognition. The work of~\cite{multi_channel_part/ChengGZWZ16} proposes to learn deep features from both global
body and local stripes to pursue better representation of human appearance. Whereas the images at hand may be not perfectly cropped (e.g., the bounding boxes are obtained by existing detection algorithms). In such a case, the fixed stripes based partition may fail. On the other hand, some other algorithms~\cite{saliency/ZhaoOW17,our_gct/Zhou18} directly perform patch-level matching, which is more flexible than
stripe-based ones, and can well address the spatial misalignment problem if patch-wise correspondences are accurately established. However, establishing dense pair-wise correspondences still remains a challenging problem.

Recently, part-based approaches are introduced into CNNs
to automatically generate semantically aligned body parts to guide feature learning. In~\cite{part_loss/YaoZZLT17}, salient body parts are generated by performing clustering on intermediate features, and an identity classification loss is imposed on both
the whole body and body parts. During testing, the generated part features are concatenated with the global feature to enhance the representative ability. Inspired by the attention model,~\cite{deep_aligned/ZhaoLZW17} proposes a part net composed
by convolutional layers to automatically detect salient body parts, and aggregate features over these parts into a global representation. Despite promising performance in handling spatial misalignment, the usage of global average pooling for feature aggregation in these approaches~\cite{part_loss/YaoZZLT17,deep_aligned/ZhaoLZW17} leads to sub-optimal results due to the ignorance of richer feature interactions and various feature significance. Part-based Convolutional Baseline (PCB) and Refined Part Pooling (RPP) have been proposed by ~\cite{sun2018beyond} to raise a new baseline and many other improved versions~\cite{sun2018beyond} have been released.

\subsection{Second Order Feature Aggregation}
Feature aggregation is an essential part of visual tasks, which means encoding and pooling of visual features to make them both efficient and discriminative, like fisher encoding~\cite{perronnin2010improving} and spatial pyramid~\cite{lazebnik2006beyond}. In CNNs, fully connected layer, average pooling and max pooling are popular operations for feature aggregation. Bilinear models are designed to separate style and content by~\cite{tenenbaum2000separating}. Then the second order pooling have been explored using deeply-learned features~\cite{lin2015bilinear,Kong_2017_CVPR} and hand-crafted features~\cite{carreira2012semantic}. However, bilinear features always involve high dimensional operations of the matrix with high computational complexity, typically on the order of a few million. Compact Bilinear Pooling~\cite{gao2016compact} provides an efficient kernelized solution with same discriminative power but with only a few thousand dimensions.

Despite being related to~\cite{deep_aligned/ZhaoLZW17}, our approach is significantly different. In this paper, we focus on improving the performance of person Re-ID by learning discriminative fine-grained feature aggregation. To this end, we present a novel WBC model. Different from~\cite{deep_aligned/ZhaoLZW17} using GAP for feature aggregation, our WBC model takes into account higher-order channel-wise feature interactions, enriching the representative ability and discriminability of the learned feature embedding. Experimental results evidence the advantages of our WBC model compared to GAP in~\cite{deep_aligned/ZhaoLZW17} for Re-ID.

\section{The Proposed Approach}
\subsection{Problem Formulation}

In this paper, we formulate the task of Re-ID as a ranking problem, where the goal is to minimize the intra-person divergence while maximize the inter-person divergence. Specifically, given an image set $\mathcal{I}=\{\mathbf{I}_1,\mathbf{I}_2,\cdots,\mathbf{I}_N\}$ with $N$ images, we form the training set into a set of triplets $\mathcal{T} = \{(\mathbf{I}_i,\mathbf{I}_j,\mathbf{I}_k)\}$, where $\mathbf{I}_i,\mathbf{I}_j,\mathbf{I}_k$ are images with identity labels $y_i$, $y_j$ and $y_k$ respectively. In a triplet unit, $(\mathbf{I}_i,\mathbf{I}_j)$ is a positive image pair of the same person (i.e., $y_i = y_j$), while $(\mathbf{I}_i,\mathbf{I}_k)$ is a negative image pair (i.e., $y_i\neq y_k$). Then the purpose of Re-ID is to rank $\mathbf{I}_j$ before $\mathbf{I}_k$ for all triplets, which can be mathematically expressed as
\begin{equation}
d(\phi(\mathbf{I}_i),\phi(\mathbf{I}_j))+ \alpha \leq d(\phi(\mathbf{I}_i),\phi(\mathbf{I}_k)) \label{eq1}
\end{equation}
where $d(\mathbf{x},\mathbf{y})=\|\mathbf{x}-\mathbf{y}\|_2$ represents the Euclidean distance, $\phi(\cdot)$ denotes the feature transformation using deep neural networks as described later, and $\alpha>0$ is the margin by which the distance between a negative image pair is greater than that between a positive image pair. To enforce this constraint, a common relaxation of Eq.~(\ref{eq1}) is the minimization of the triplet hinge loss as
\begin{equation}
\ell_{tri}(\mathbf{I}_i,\mathbf{I}_j,\mathbf{I}_k)=\Big[ d(\phi(\mathbf{I}_i),\phi(\mathbf{I}_j)) - d(\phi(\mathbf{I}_i),\phi(\mathbf{I}_k)) + \alpha \Big]_{+}
\label{eq2}
\end{equation}
where the operator $[\cdot]_{+}=\mathrm{max}(0,\cdot)$ represents the hinge loss. The whole loss function for all triplets in training set is then expressed as
\begin{equation}
\mathcal{L}(\phi) = \frac{1}{|\mathcal{T}|} \sum_{(\mathbf{I}_i,\mathbf{I}_j,\mathbf{I}_k) \in \mathcal{T}} \ell_{tri}(\mathbf{I}_i,\mathbf{I}_j,\mathbf{I}_k)
\label{eq3}
\end{equation}
where $|\mathcal{T}|$ denotes the number of triplets in $\mathcal{T}$. The loss function will be presented in section 3.4.
\subsection{Salient Part-based Representation}
\label{sec:salient_part}
\begin{figure}[!t]
	\centering
	\includegraphics[width=\linewidth]{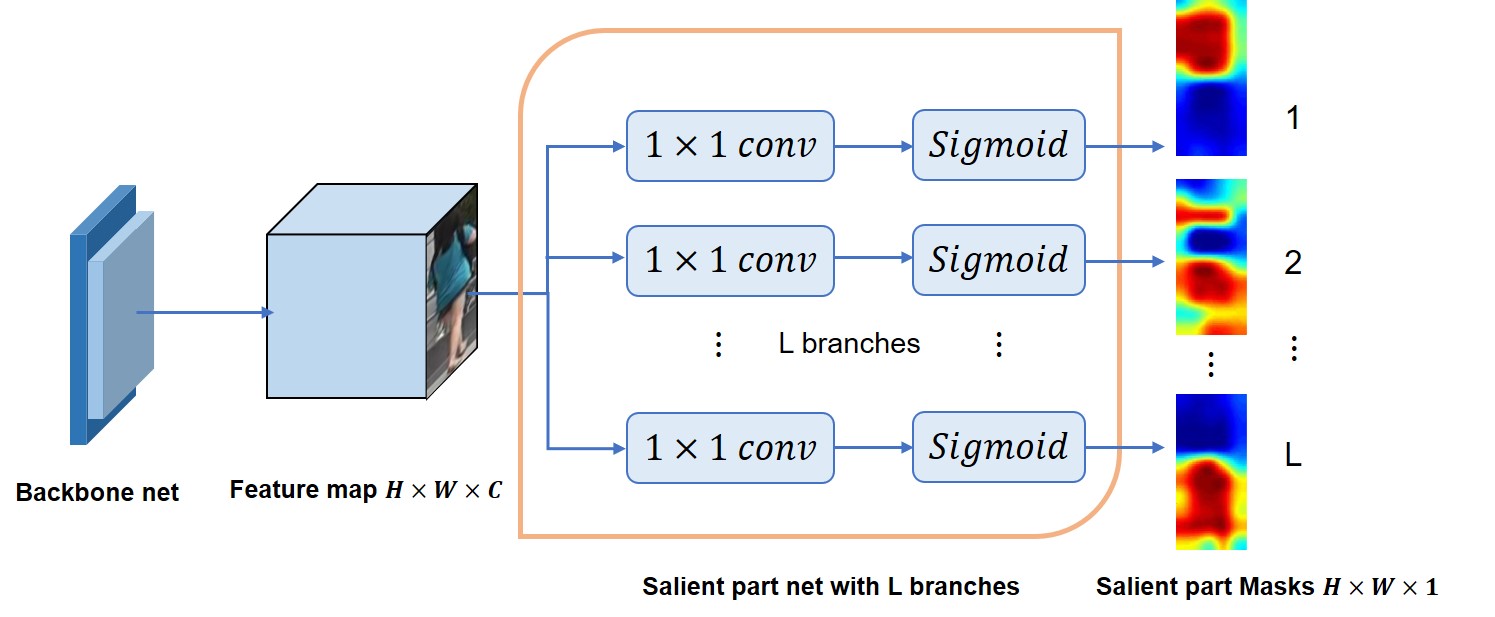}
	\captionsetup{font={footnotesize}}
	\vspace{-1ex}
	\caption{Illustration of the architecture of salient part net.}\label{fig:part_net}
	\vspace{-2ex}
\end{figure}
For Re-ID, one of the most technical bottleneck is spatial misalignment caused by variations in views and human poses. Local representations are computed typically by partitioning full images into fixed horizontal stripes or grids, assuming the bounding boxes are well detected and spatial distributions of human body are similar. To deal with this issue, we adopt a salient part-based representation (spatial attention module) to measure the importance of human appearance parts in spatial domain. We call it salient part net. Images are passed into the backbone to get the intermediate feature maps. The salient part masks are generated by feeding the obtained feature maps into the salient part net. Specifically, the salient part nets consist of $L$ branches, each corresponding to a certain part of the human body. For simplicity, each branch is composed of a $1\times 1$ conv, followed by a Sigmoid layer to map the values to $(0, 1)$. So the representations over the parts are learned end-to-end and the similarities between the corresponding parts are aggregated. Therefore, each convolutional branch can learn attention mask of a local area of original feature map due to feature concatenation and triplet loss learning. 



Figure \ref{fig:part_net} illustrates the architecture of the salient part net. In specific, the salient part net consists of $L$ branches, and each branch is composed of a $1\times{1}$ convolutional layer followed by a nonlinear sigmoid layer. The input to the salient part net is the 3-dimension intermediate convolutional feature maps, and its outputs are $L$2-dimension salient part masks. Specifically, let $\mathbf{F} \in \mathbb{R}^{H\times{W}\times{C}}$ represent the input feature maps to the salient part net, then we can estimate the part masks $\mathbf{M}_l \in \mathbb{R}^{H\times{W}\times{1}}, l \in \{1,\cdots,L\}$ as
\begin{equation}
\mathbf{M}_l = \Phi_{\mathrm{SalientMask}_{l}}(\mathbf{F})
\label{eq:mask}
\end{equation}
where $\Phi_{\mathrm{SalientMask}_{l}}(\cdot)$ represents the $l^{\mathrm{th}}$ salient part mask generator. In Eq.~(\ref{eq:mask}), the values of elements in each $\mathbf{M}_l$ are within the range $(0,1)$, reflecting the relative importance of their corresponding local features. Taking $\mathbf{M}_l$ as the automatically learned weights, we can compute the part-based feature $\mathbf{F}_{l}$ using the proposed WBC as
\begin{equation}
\mathbf{F}_l = \Psi_{\mathrm{WBC}}(\mathbf{M}_l, \mathbf{F})
\end{equation}
where $\Psi_{\mathrm{WBC}}(\cdot, \cdot)$ represents the proposed feature coding algorithm and will be discussed later. It is worth noting that different from existing part-aligned Re-ID method~\cite{deep_aligned/ZhaoLZW17} which uses global average pooling for feature aggregation, our WBC is able to fully explore richer higher-order channel-wise feature interactions, improving the representative ability and discriminability of feature aggregation.

Afterwards, the encoded feature of each part $\mathbf{F}_l$ is passed into a linear embedding for dimension reduction. Let $\mathbf{F}_l'$ denote the dimension-reduced feature of $\mathbf{F}_l$, then the discriminative part-aligned feature representation is formed by concatenating $\mathbf{F}_l'$ for each part, followed by $\ell_2$ normalization,
\begin{equation}
\mathbf{f} =\phi(\mathbf{I}) = \Big\|[(\mathbf{F}_1')^{\top},(\mathbf{F}_2')^{\top},\cdots,(\mathbf{F}_L')^{\top}]^{\top}\Big\|_2
\end{equation}
The obtained feature representation $\mathbf{f}$ is then utilized as the feature transformation $\phi(\mathbf{I})$ in Eq. (\ref{eq1}).

\begin{figure}[!t]
	\centering
	\includegraphics[width=\linewidth]{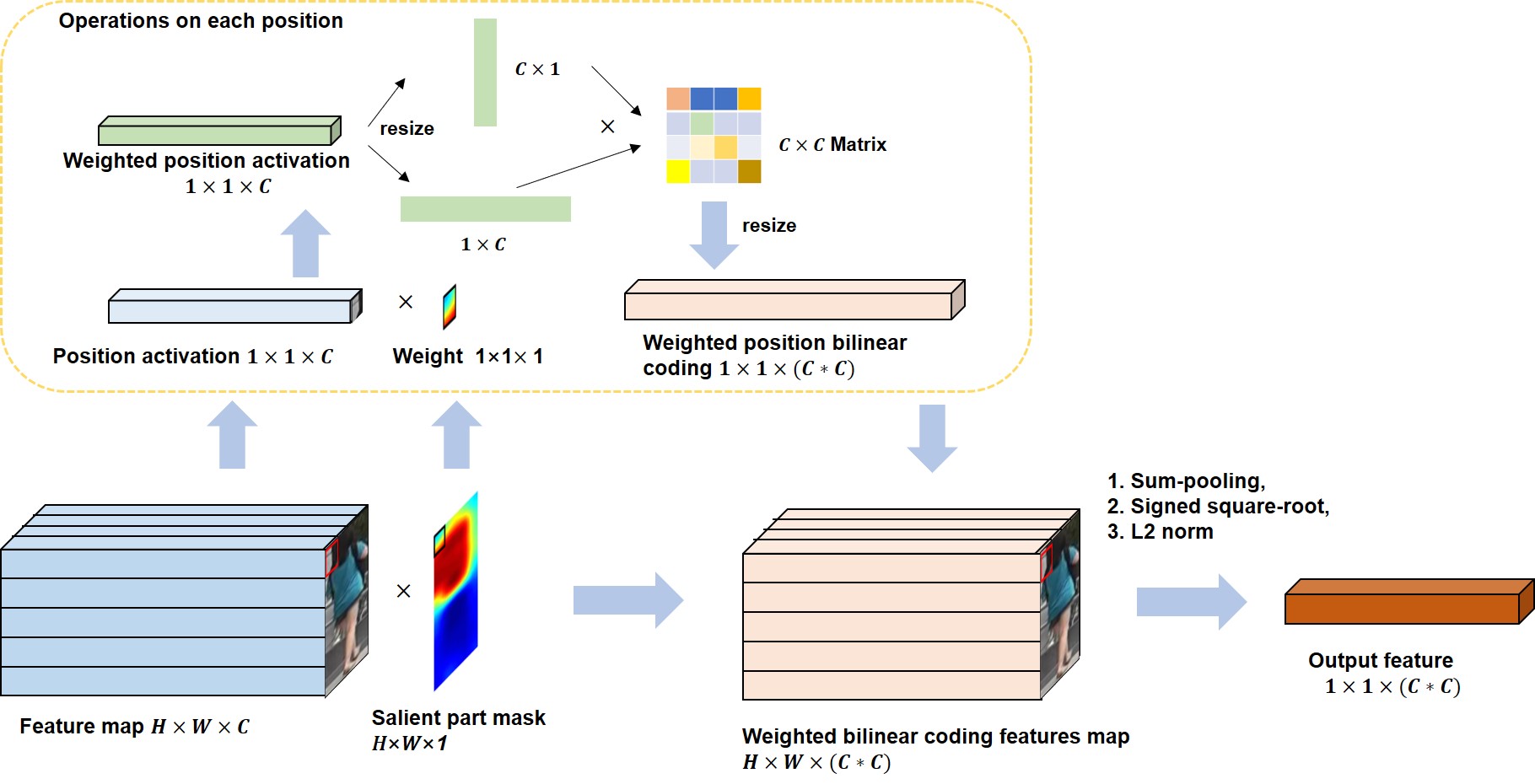}
	\captionsetup{font={footnotesize}}
	\vspace{-1ex}
	\caption{Illustration of the WBC model applied on one salient part mask.}\label{fig:wbc}
	\vspace{-2ex}
\end{figure}

\subsection{Weighted Bilinear Coding}
\label{sec:wbc}
Given the input feature maps $\mathbf{F} \in \mathbb{R}^{H\times{W}\times{C}}$, much identity-aware discriminative information of the input image $\mathbf{I}$ is implicitly captured. However, how to aggregate the local features of $\mathbf{F}$ to fully explore its representative and discriminative potential for Re-ID remains a problem. Most of the existing algorithms~\cite{deep_aligned/ZhaoLZW17,part_loss/YaoZZLT17} adopt global average pooling (GAP) for the sake of efficiency and simplicity. However, GAP only captures the first-order statistics of local features and considers all the units inside the feature maps equally important. This may undermine both the representative and discriminative ability of the final representation.
 Bilinear coding~\cite{lin2015bilinear, Kong_2017_CVPR} is recently introduced into the CNN network to model the higher-order channel-wise feature interactions, enhancing the representative ability of the learned deep features. Originally, the bilinear coding takes all the local features as input and outputs a representation $\mathbf{B}$ as follows:
\begin{equation}
\mathbf{B} = \sum_{p=1}^H\sum_{q=1}^W \mathbf{F}(p,q)^T\mathbf{F}(p,q)
\label{eq:raw_formulation}
\end{equation}
where $\mathbf{F}(p,q) \in \mathbb{R}^{1 \times C}$ is the local feature at the $(p,q)$-$th$ location. Nevertheless, it is suboptimal for Re-ID without considering the various impacts of different local features.

To address the above-mentioned issue, we introduce a novel \emph{weighted bilinear coding} model to adaptively weigh local features at different locations according to their relative importance. The process is fully shown in Figure \ref{fig:wbc}. In our approach, the relative importance is automatically captured in the salient part masks generated by the salient part net. And the weighted bilinear coded feature is calculated as
\begin{equation}\small
\Psi_{\mathrm{WBC}}(\mathbf{M}_l, \mathbf{F})= \sum_{p=1}^H\sum_{q=1}^W (\mathbf{M}_l(p,q) \mathbf{F}(p,q))^T(\mathbf{M}_l(p,q)\mathbf{F}(p,q))
\label{eq:wbc_formulation}
\end{equation}
where $\mathbf{M}_l$ is the $l$-$th$ part mask generated from Eq.~(\ref{eq:mask}). 

The details of bilinear operations on each position of activation map are illustrated in Figure \ref{fig:wbc}. The salient part masks are generated by salient part net. For each salient mask, the weighted activation $1 \times 1 \times C$ are resized to $C \times 1$ and $1 \times C$, then they do the outer product operation to produce a $C \times C$ matrix in Eq.~(\ref{eq:wbc_formulation}) . In this way, local features are weighted adaptively such that more critical units can play a more important role \sout{in subsequent recognition process}.The part feature matrix $\mathbf{F}_l = \Psi_{\mathrm{WBC}}(\mathbf{M}_l, \mathbf{F})$ is then reshaped to a $C^2$ length vector and further passed through a signed square-root step ($\mathbf{F}_l = sign(\mathbf{F}_l)\sqrt{|\mathbf{F}_l|}$) and a $L2$ normalization layer before fed into the linear embedding layer to perform feature dimension reduction. In the  WBC model, the outer product helps to capture richer local feature interactions, enhancing the representative ability of the deep features. Meanwhile, the weighting scheme encodes the relative importance of different local features, leading to more discriminative representation.

\subsection{Loss Function}
\label{sec:loss}

We use the batch hard triplet loss function which was originally proposed in~\cite{hermans2017defense}. To form a batch, we randomly sample $P$ identities and randomly sample $K$ clips for each identity (each clip contains $T$ frames); Totally there are $PK$ clips in a batch. For each sample $a$ in the batch, the hardest positive and the hardest negative samples within the batch are selected when forming the triplets for computing the loss $L_{triplet}$.
\begin{equation}
L_{tripletloss} = \sum_{i=1}^{P}\sum_{a=1}^{K}[m + \overbrace{\underset{p=1...k}{max}D(\phi(\mathbf{I}_a^i),\phi(\mathbf{I}_p^i))}^{hardest\ positive} - \underbrace{\underset{\underset{\underset{j \neq i}{n=1...K}}{j=1...P}}{min}D(\phi(\mathbf{I}_a^i),\phi(\mathbf{I}_n^j))}_{hardest\ negative}]
\label{eq:loss}
\end{equation}

\subsection{Flexibility and portability}
Functioning as a feature aggregation part, the proposed WBC module can be readily combined with state-of-the-art feature networks to enhance their recognition performance. In this section, we combine WBC with the popular PCB-RPP~\cite{sun2018beyond} method to further boost its performance.  In PCB-RPP~\cite{sun2018beyond}, it used a learned classifiers containing a linear layer followed by Softmax activation to relocate and weigh the predicted probability of position activation in feature map belonging to local part. In specific, we take the output after the Softmax operation in RPP module in [46] as the salient mask, then features of each part are encoded by the proposed WBC module to form the final part features. For fair comparison, other parts of the PCB-RPP algorithm keep exactly the same. This part of experiments will be discussed in detail in section 4.


\section{Experiments}

In this section, we describe our evaluation protocols and provide a detailed ablation study of the proposed architecture. Extensive experiments on three challenging benchmarks including Market-1501~\cite{market_dataset/ZhengSTWWT15}, DukeMTMC-reID~\cite{duke_dataset/RistaniSZCT16} and CUHK03~\cite{cuhk03_dataset/LiZXW14} show that the proposed algorithm performs favorably against other  approaches.

\subsection{Datasets and evaluation metric}

\noindent
{\bf Market-1501} is one of the most challenging datasets for Re-ID. It is collected in front of a supermarket
using five high-resolution and one low-resolution cameras. In total, this dataset contains 32,768 annotated bounding boxes belonging to 1,501 identities obtained from existing pedestrian detection algorithm~\cite{felzenszwalb2010object}.
Among the 1,501 identities, 750 individuals are set for training and the rest for testing.

\noindent
{\bf DukeMTMC-reID} consists of 36,411 bounding boxes with labeled IDs, among which 1,404 identities appear in more than two cameras and 408 identities (distractor ID appears in only one camera). This dataset is further divided into training subset with 16,522 images of 702 identities, and testing subset with 2228 query images of the other 702 identities and 17,661 gallery images (images of the remaining 702 IDs and 408 distractor IDs).

\noindent
{\bf CUHK03} contains 13,164 images of total 1,360 persons captured under six cameras. In this dataset, each individual appears in two disjoint camera views, and on average 4.8 images of each view are collected for each person. The performance is originally evaluated on 20 random splits of 1276 persons for training and 100 individuals for testing, which is time-consuming. Instead, we follow the evaluation protocol in~\cite{re-ranking/ZhongZCL17} to split the dataset into training set composed of 767 identities and testing set with the rest identities. The CUHK03 benchmark provides both hand-labeled and DPM-detected~\cite{felzenszwalb2010object} bounding boxes, we conduct experiments on both of them to validate the effectiveness of the proposed algorithm.

Following recent literature, all experiments are evaluated under the single-shot setting, where a ranking score is generated for each query image and all the scores are averaged to get the final recognition accuracy. The recognition performance are evaluated by the cumulative matching characteristic (CMC) curve and the mean average precision (mAP) criterion. The CMC curve represents the expected probability of finding the first correct match for a probe image in the top $r$ match in the gallery list. And as supplementary, mean average precision summarizes the ranking results for all the correct matches in the gallery list.

\subsection{Implementation details}

The proposed algorithm is implemented using PyTorch~\cite{caffe/JiaSDKLGGD14} on two NVIDIA GTX 1080 GPUs with 8GB memory. We adopt the GoogLeNet~\cite{googlenet/SzegedyLJSRAEVR15} and Resnet-50~\cite{resnet/HeZRS16} as the backbones CNN network. 

\subsubsection{GoogLeNet Backbone}
Feature maps are extracted from the $inception\_4e$ layer when GoogLeNet, followed by a $1\times 1$ convolutional layer with $512$ feature channels. The input images are resized to $160 \times 80$. The number of parts ($L$) generated by the salient part net is discussed later, and distance margin $\alpha$ in Eq.~(\ref{eq:linear}) is set to $0.3$ throughout the experiments. The whole network is optimized using stochastic gradient descent (SGD) method on mini-batches. Each mini-batch is sampled with randomly selected P identities and randomly sampled K images for each identity from the training set. $P=32, K=4$ when Market-1501 and DukeMTMC-reID, and $P=16, K=4$ when CUHK03. The initial learning rate is set to 0.008, and it is divided by 2 every 4,000 iterations. The weight decay and the momentum are set to 0.0005 and 0.9.
The channel dimension is set to 1024 after the linear embedding in Weighted Bilinear Coding Module.
\subsubsection{Resnet-50 Backbone}
We also implement a Resnet-50 backbone as comparison. The Resnet-50 baseli Weights of pretrained ResNet-50 \cite{resnet/HeZRS16} initializes the backbone. We removed the last spatial downsampling in the last residual block in ResNet50 to increase the size of feature map. Input images are resized to 384$\times$128, so the final output feature map size before pooling layers is 24$\times$8 and the channel dimension is 2048. ADAM optimizer is used with momentum 0.9. We set 1.2 for the margin parameter for triplet loss and $10^{-4}$ for initial learning rate. Learning rate decays to $10^{-5}$ and $10^{-6}$ after training for 120 and 160 epochs. Random horizontal flipping for data augmentation and features averaged from original images and the horizontally flipped versions for evaluation are deployed. The channel dimension is set to 1024 after the linear embedding in Weighted Bilinear Coding Module.

\subsection{Ablation study}

\subsubsection{Modules validity analysis}

To further validate the proposed Re-ID algorithm, we conduct experiments on several baselines and compare with them on GoogLeNet Backbone. In specific, we develop three baselines including GAP Net, GAP+SalientPart Net and BC Net on GoogLeNet Baseline as follows.

{\bf GAP Net} is implemented by removing the salient part net from our method and replacing the proposed WBC with global average pooling, and other settings are kept exactly the same.

{\bf GAP+SalientPart Net} is implemented by substituting the proposed WBC with global average pooling, and other settings are kept exactly the same.  $L$ is set to 3 in this part.

{\bf BC Net} is implemented by removing the salient part net from our method and replacing the proposed WBC with original bilinear coding (BC), and other settings are kept exactly the same.

\begin{figure*}[!t]
	\centering
	\begin{tabular}{@{}C{4.3cm}@{}C{4.3cm}@{}C{4.3cm}@{}}
		\includegraphics[width=4.2cm]{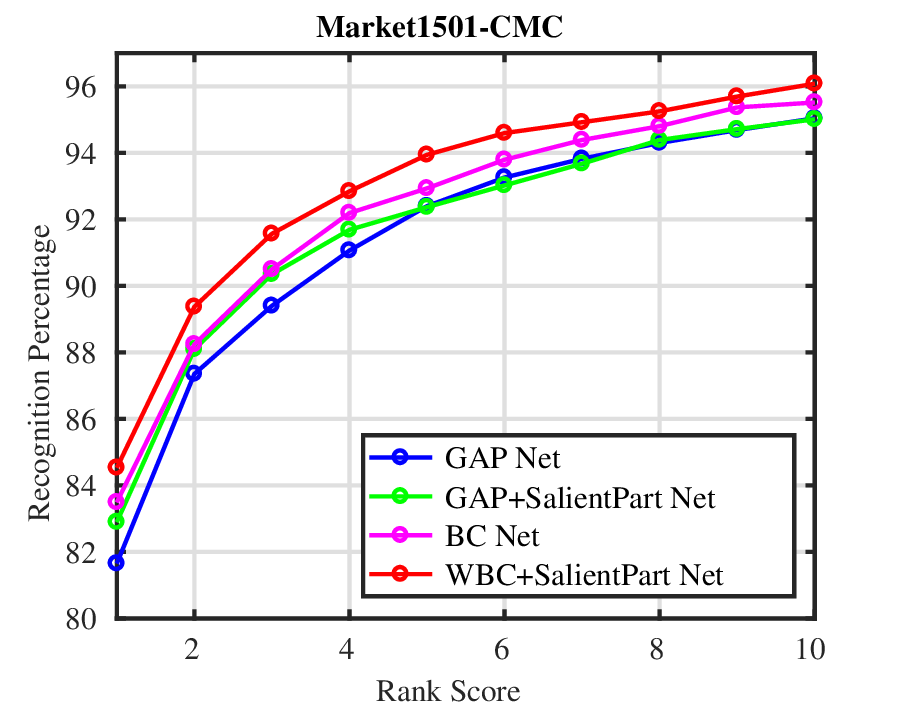}&\includegraphics[width=4.2cm]{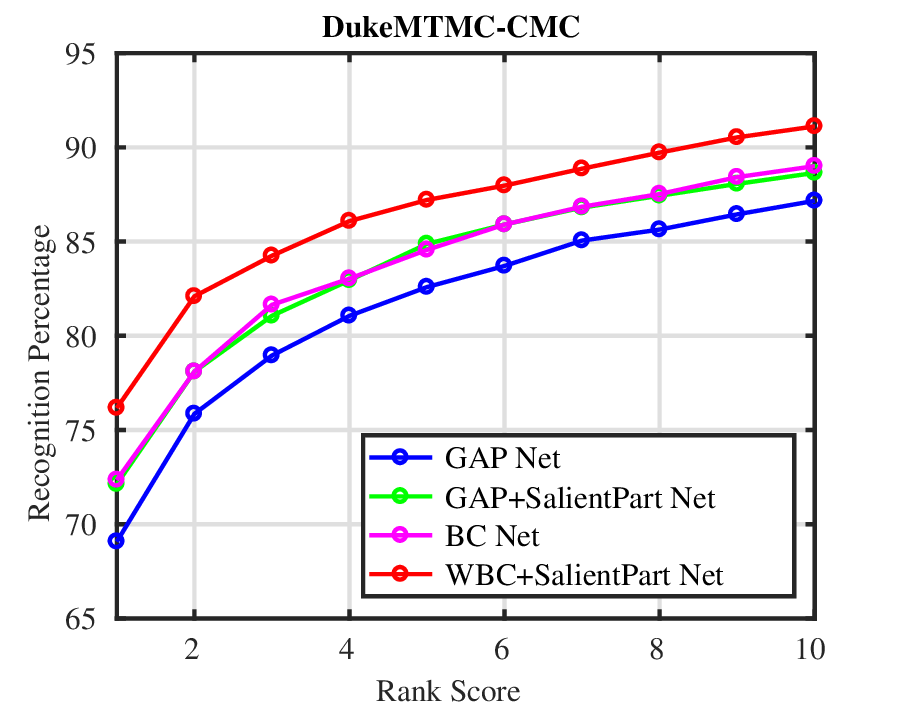}&
		\includegraphics[width=4.2cm]{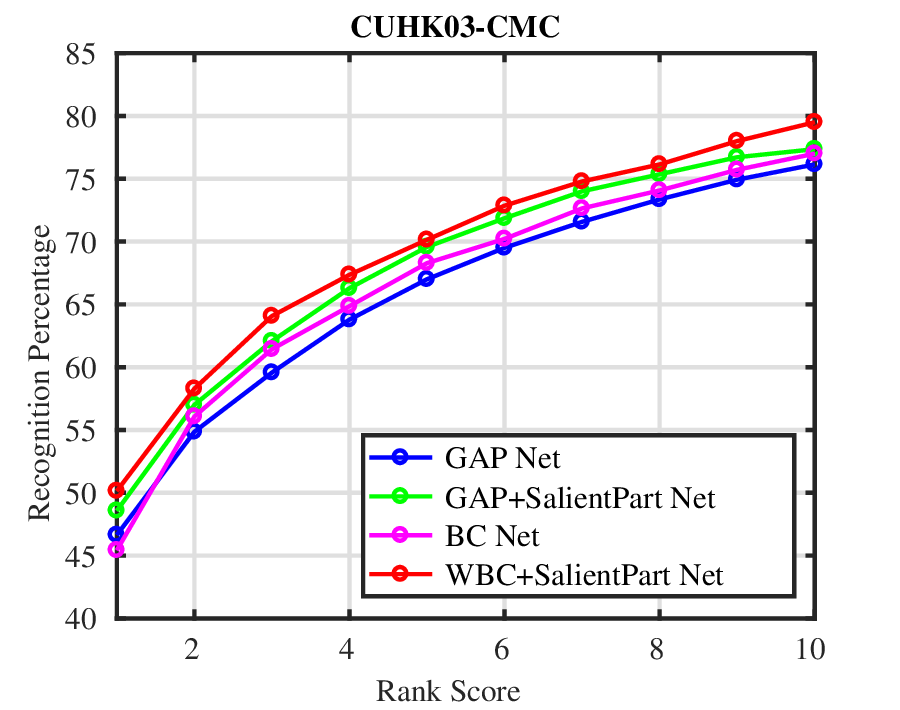}\\
		\small{(a) Market1501} & \small{(b) DukeMTMC-reID} &\small{(c) CUHK03}  \\
	\end{tabular}
	\vspace{-2ex}
	\caption{Comparison results of the proposed algorithm with three baselines in terms of the top $r$ matching rate using CMC. Best viewed in color.}\label{fig:ablative_comparison}
\end{figure*}

Our method is referred to as {\bf WBC+SalientPart Net}.  $L$ is set to 3 in this part. The recognition performance of each network on three challenging large-scale person re-identification benchmarks are shown in Figure~\ref{fig:ablative_comparison}.

As demonstrated in Figure~\ref{fig:ablative_comparison}, our {\bf WBC+SalientPart Net} performs consistently better than the other three baselines on all the three benchmarks. More specifically, in comparison with GAP, our approach has 2.8 \%, 4.1 \% and 4.7 \% rank-1 performance gain on the Market1501 dataset, DukeMTMC-reID dataset and CUHK03 dataset respectively. Please note here the {\bf GAP+SalientPart Net} baseline also aggregates local features over each salient part and concatenate them to form the final representation, but our approach achieves better performance, validating the more powerful representative ability of our \emph{weighted bilinear coding} (WBC) than GAP. Furthermore, the comparison results with the {\bf BC Net} baseline demonstrate the effectiveness of the weighting scheme in our WBC framwork.

\begin{figure}[!t]
  \centering
  \includegraphics[width=\linewidth]{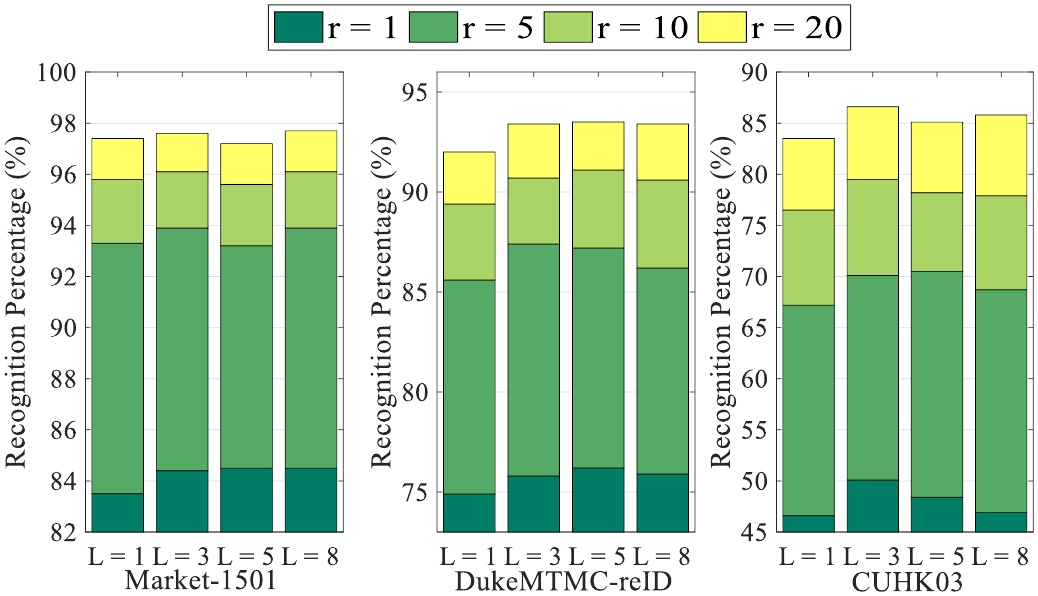}
  \caption{Recognition performance of different number of salient parts. As shown, the number of parts set to $L=3$ is a good compromise between efficiency and effectiveness.  }
  \vspace{-2ex}
\label{fig:part_num}
\end{figure}
\renewcommand\arraystretch{1.2}
\begin{table}[!t]\small
\caption{Analysis on the influence of different number of salient parts, both the CMC (\%) top $r$ ranking rates and the mAP (\%) are reported.
\label{tab:number_of_parts}}
\vspace{-2ex}
\begin{center}
\begin{tabular}{c| c| ccccc}
\hline
Datasets& \# parts  & r=1 & r=5 & r=10 & r=20& mAP \\
 \hline\hline
 & L = 1&  $83.5$ & $93.3$ & $95.8$ & $97.4$ & 66.67\\
Market1501& L = 3 &84.4 &  $93.9$ & $96.1$ & $97.6$ & $67.61$\\

&L = 5 &$84.5$ & $93.2$ & $95.6$ & $97.2$ &$67.07$\\
&L = 8&$84.5$ & $93.9$ & $96.1$ & $97.7$ &$68.69$\\
\hline\hline
&L = 1&  $74.9$ & $85.6$ & $89.4$ & $92.0$ & $54.75$\\
DukeMTMC &L = 3 &  $75.8$ & $87.4$ & $90.7$ & $93.4$ & $56.94$\\

&L = 5 &$76.2$ & $87.2$ & $91.1$ & $93.5$ &$56.85$\\
&L = 8&$75.9$ & $86.2$ & $90.6$ & $93.4$ &$56.40$\\

\hline\hline
&L = 1&  $46.6$ & $67.2$ & $76.5$ & $83.5$& $44.3$\\
CUHK03 &L = 3 &  $50.1$ & $70.1$ & $79.5$ & $86.6$ & $47.72$\\
&L = 5 &$48.4$ & $70.5$ & $78.2$ & $85.1$ &$46.51$\\
&L = 8& $46.9$ & $68.7$ & $77.9$ &$85.8$ & $44.65$\\
\hline
\end{tabular}
\end{center}
\vspace{-3ex}
\end{table}
\subsubsection{ Analysis on different number of salient parts}

We empirically study the optimal number $L$ of salient parts on each dataset. In specific, we record the recognition performance of $L=1,3,5,8$, respectively. As shown in Table~\ref{tab:number_of_parts} and Figure~\ref{fig:part_num}, on the Market1501, the best result is obtained with $L=8$, which outperforms $L=1$ by 1\% in CMC recognition rate ($r=1$) and 2\% in term of mean average precision (mAP). On the DukeMTMC-reID dataset, $L = 3$ and $L = 5$ achieve better results than $L=1, 8$, where $L=5$ achieves the best CMC performance (76.2\% for $r=1$), while $L=3$ has the highest mAP (56.94\%). On the CUHK03 (with human-labeled bounding boxes), $L=3$ outperforms all other settings ($L = 1, 5, 8$) in terms of both CMC and mAP. Overall, on all three datasets, multiple salient parts (with $L$ bigger than 1) indeed bring performance gain to the proposed model. Since using multiple parts requires more computational resources, $L$ is suggested to be set to 3 for the balance between effectiveness and efficiency. In the following part, the best result reported on each dataset is used for comparison with other state-of-the-art algorithms without special clarification.

\subsubsection{Performance comparisons between different backbones and Generalization verification analysis}
In this section, Firstly we compare the performances of two backbone networks, the purpose of comparison is to measure the performance gain when transferring to the Resnet-50. Because the Resnet-50 is also the backbone network of PCB-RPP~\cite{sun2018beyond} many other methods, we can more easily and directly compare with others algorithms. Secondly, we assemble our WBC framework with PCB+RPP method mentioned above in section 3.5.

\begin{table}[th]
\begin{center}
\caption{
Performance comparisons between different backbone on datasets Market1501 and DukeMTMC-reID
}
\label{table2}
\resizebox{12cm}{!}{
\begin{tabular}{l|ccc|ccc}
\hline
\multirow{2}{*}{Method} &  \multicolumn{3}{c|}{Market1501} & \multicolumn{3}{c}{DukeMTMC} \\
\cline{2-7}          & mAP & rank1 & rank5  & mAP & rank1 & rank5\\
\hline \hline
\textbf{GAP-GoogLeNet}   &59.6 &81.8 &92.3 &47.2 &69.3 &82.4\\
\textbf{GAP-Resnet}   &66.0 &84.6 &91.1 &55.3 &72.3 &86.5\\
\textbf{BC-GoogLeNet}  &63.4 &83.7 &88.3 &53.9 &72.6 &84.9\\
\textbf{BC-Resnet}  &77.1  &87.4 &94.0 &63.2 &77.5 &89.1 \\
\textbf{WBC-full-GoogLeNet}  &67.6 &84.4 &93.9 &56.9 &75.8 &87.4 \\
    \textbf{WBC-full-Resnet}  &\textbf{81.9} &\textbf{92.1} &\textbf{96.5} &\textbf{69.1} &\textbf{81.7} &\textbf{91.4}\\
\hline \hline
\textbf{PCB-triplet} &74.9 &88.9 &94.7 &64.1 &77.6 &88.9 \\
\textbf{PCB\cite{sun2018beyond}} &77.4 &92.3 &97.2 &66.1 &81.8 &89.4\\
\textbf{PCB+RPP\cite{sun2018beyond}} &81.6 &\textbf{93.7} &97.5 &69.2 &83.3 &90.5\\
    \textbf{WBC+RPP} &\textbf{83.3} &93.6 &\textbf{98.1} &\textbf{72.7} &\textbf{84.1} &\textbf{91.7}\\

\hline
\end{tabular}}
\end{center}
\vspace{-0.4cm}
\end{table}

{\bf GAP-GoogLeNet} is the GAP Net in section 4.3.1 based on GoogLeNet backbone. {\bf GAP-Resnet} is the same structure based on Resnet-50 backbone. {\bf BC-GoogLeNet} is the BC Net in section 4.3.1 based on GoogLeNet backbone removing the salient part net from our method and replacing the proposed WBC with original bilinear coding (BC). {\bf BC-Resnet} is the same structure based on Resnet-50 backbone. {\bf WBC-full-GoogLeNet} is the complete implementation as {\bf WBC+SalientPart Net} in section 4.3.1 based on GoogLeNet backbone. {\bf WBC-full-Resnet} is the same structure based on Resnet-50 backbone.  $L$ is set to 3 in this part. 

The comparison of different backbones performance is shown in the upper part of Table~\ref{table2}. After transferring to Resnet-50, our method {\bf WBC+SalientPart Net} has significant performance gain on Market1501 and DukeMTMC datasets. Rank1 accuracy increased by 7.7 \% and 5.9 \% while mAP rises by 14.3 \% and 12.2 \% on both datasets. Incomplete implements such as GAP Net and BC Net can also confirm that considerable gain has been achieved after transferring to Resnet-50 backbone network.

The PCB-RPP~\cite{sun2018beyond} model uses Cross Entropy Loss as loss function, we completely follow the setting of the paper. The WBC module serves as the pooling operation upon the weighted part-based features. The part-based feature channel dimension is set to 2048 after the Weighted Bilinear Coding layer. The other settings of RPP remain the same. As shown in the lower part of Table~\ref{table2}, the mAP increased by 1.7 \% and 3.5 \% on both datasets. Apart from rank1 accuracy in Market1501, Other performance indicators listed have improved significantly, which effectively prove that our framework can be flexibly plugged into existing deep architectures like PCB-RPP.

\subsection{Comparison with state-of-the-arts}

\renewcommand\arraystretch{1.2}
\begin{table}[!htbp]\small
\caption{Comparison results of top $r$ matching rate using CMC (\%) and mean average precision (mAP \%) on the Market1501 dataset.
\label{tab:market_Dataset}}
\vspace{-2ex}
\begin{center}
\begin{tabular}{r|ccccc}
\hline
 Methods  & r=1 & r=5 & r=10 & r=20 &mAP\\
 \hline\hline
LOMO+XQDA~\cite{lomo_xqda/LiaoHZL15}  &  $43.8$ & $-$ & $-$ & $-$ & 22.2\\
BoW~\cite{market_dataset/ZhengSTWWT15}  &  $44.4$ & $63.9$ & $72.2$ & $79.0$ & 20.8\\

WARCA~\cite{warca/JoseF16}  & $45.2$ & $68.2$ & $76.0$ & $-$ & $-$\\
SCSP~\cite{scsp/ChenYCZ16} &$51.9$ & $-$ & $-$ & $-$ & $26.4$\\
Re-ranking~\cite{re-ranking/ZhongZCL17} &  $77.1$ & $-$ & $-$ & $-$ & $63.6$\\
DNS~\cite{dns/ZhangXG16} &$55.4$ & $-$ & $-$ & $-$ & $29.9$\\

Gated S-CNN~\cite{gated_siamese/VariorHW16}&$65.9$ & $-$ & $-$ & $-$ & $39.6$\\
P2S~\cite{P2S/ZhouWWGZ17} &$70.7$ & $-$ & $-$ & $-$ & $44.2$\\
CADL~\cite{consistent_aware/LinRLFZ17} &$73.8$ & $-$ & $-$ & $-$ & $47.1$\\
Spindle Net~\cite{spindle_net/ZhaoTSSYYWT17}  &$76.9$ & $91.5$ & $94.6$ & $96.7$ &\\
LSRO~\cite{LSRO/ZhengZY17} &$79.3$ & $-$ & $-$ & $-$ & $56.0$\\
MSCAN~\cite{deep_context/LiC0H17} &$80.3$ & $-$ & $-$ & $-$ &$57.5$\\
PADF~\cite{deep_aligned/ZhaoLZW17}  &$81.0$ & $92.0$ & $94.7$ & $-$&$63.4$\\
SSM~\cite{SSM/BaiBT17} &$82.2$ & $-$ & $-$ & $-$ & $68.8$\\
SVDNet~\cite{SVDNET/SunZDW17} &$82.3$ & $92.3$ & $ 95.2$ & $-$ & $62.1$\\
ACRN~\cite{ACRN/SchumannS17} &$83.6$ & $92.6$ & $95.3$ & $97.0$ &$62.6$\\
PDC~\cite{pose_driven/SuLZX0T17} &$84.1$ & $f92.7$ & $94.9$ & $96.8$ & $63.4$\\
JLML~\cite{JLML/LiZG17} &$85.1$ & $-$ & $-$ & $-$ & $65.5$\\
AOS \cite{huang2018adversarially} &$86.5$ & $-$ & $-$ & $-$ &$70.4$\\
MLFN \cite{DBLP:journals/corr/abs-1803-09132} &$90.0$ & $-$ & $-$ & $-$ &$74.3$\\
HA-CNN \cite{li2018harmonious} &$91.2$ & $-$ & $-$ & $-$ &$75.7$ \\
PCB-RPP~\cite{sun2018beyond} &$93.8$ & $97.5$ & $98.5$ & $-$ & $81.6$\\

 \hline\hline
\textbf{WBC-full-GoogLeNet} &  $84.5$ & $93.9$ & $96.1$ & $97.7$ &$68.7$\\
\textbf{WBC-full-Resnet} &  $92.1$ & $96.5$ & $98.6$ & $99.7$ &$81.9$\\

\hline
\end{tabular}
\vspace{-3ex}
\end{center}
\end{table}

In this section, we present the comparison results with state-of-the-art algorithms on Market-1501, DukeMTMC-reID and CUHK03 benchmarks.
\vspace{-1ex}
\subsubsection{ Results on Market1501}

On the Market1501 dataset, we compare the proposed Re-ID algorithm based on two backbones with many Re-ID algorithms, including feature designing based algorithms: LOMO+XQDA~\cite{lomo_xqda/LiaoHZL15} and BoW~\cite{market_dataset/ZhengSTWWT15}; metric learning based algorithms:
weighted approximate rank component analysis (WARCA)~\cite{warca/JoseF16}, SCSP~\cite{scsp/ChenYCZ16}, Re-ranking~\cite{re-ranking/ZhongZCL17} and DNS~\cite{dns/ZhangXG16}; and deep learning based algorithms: Gated S-CNN~\cite{gated_siamese/VariorHW16}, set similarity learning (P2S)~\cite{P2S/ZhouWWGZ17}, consistent aware deep network (CADL)~\cite{consistent_aware/LinRLFZ17}, Spindle Net~\cite{spindle_net/ZhaoTSSYYWT17}, LSRO~\cite{LSRO/ZhengZY17}, multi-scale context aware network (MSCAN)~\cite{deep_context/LiC0H17}, part aligned deep features (PADF)~\cite{deep_aligned/ZhaoLZW17}, SSM~\cite{SSM/BaiBT17}, SVDNet~\cite{SVDNET/SunZDW17}, ACRN~\cite{ACRN/SchumannS17}, JLML~\cite{JLML/LiZG17}, pose-driven deep convolutional model (PDC)~\cite{pose_driven/SuLZX0T17}, Harmonious Attention Network (HA-CNN)~\cite{li2018harmonious}, AOS~\cite{huang2018adversarially}, Multi-Level Factorisation Net (MLFN)~\cite{DBLP:journals/corr/abs-1803-09132} and PCB-RPP~\cite{sun2018beyond}.

The detailed comparison results are reported in Table~\ref{tab:market_Dataset}, from which we can see that in general our approach based on Resnet backbones outperforms most of other state-of-the-art algorithms except a slightly lower performance than PCB-RPP~\cite{sun2018beyond}. But it has been proven that our framework with RPP in Table~\ref{table2} has better performance than PCB-RPP.

Only using GoogLeNet as backbone, its performance can beat other methods listed above JLML~\cite{JLML/LiZG17}. Considering that JLML~\cite{JLML/LiZG17} utilizes the ResNet-50~\cite{resnet/HeZRS16} as the backbone network, which is more powerful than our adopted GoogLeNet, our performance is still competitive. Besides, our approach achieves very competitive recalls (the second best mAP). It is worth noting that PADF~\cite{deep_aligned/ZhaoLZW17} and PDC~\cite{pose_driven/SuLZX0T17} are two deep learning based methods which utilize part-based strategy and adopt global average pooling for feature aggregation. In comparison to PADF and PDC, the proposed model consistently generates better performance, demonstrating the superiority of our WBC model over global average pooling.

\subsubsection{Results on DukeMTMC-reID}
\renewcommand\arraystretch{1.2}
\begin{table}[!htb]\small
	\caption{Comparison results of top $r$ matching rate using CMC (\%) and mean average precision (mAP \%) on the DukeMTMC-reID dataset. 
		\label{tab:duke_Dataset}}
	\vspace{-2ex}
	\begin{center}
		\begin{tabular}{r|ccccc}
			\hline
			Methods  & r=1 & r=5 & r=10 & r=20& mAP \\
			\hline\hline
			LOMO+XQDA~\cite{lomo_xqda/LiaoHZL15} &  $52.4$ & $74.5$ & $83.7$ & $89.9$\\
			BoW~\cite{market_dataset/ZhengSTWWT15}  &  $25.1$ & $-$ & $-$ & $-$ & $12.2$\\
		
			LSRO~\cite{LSRO/ZhengZY17}  &$67.7$ & $-$ & $-$ & $-$ &$47.1$\\
			ACRN~\cite{ACRN/SchumannS17} &$72.6$ & $84.8$ & $88.9$ & $91.5$ &$52.0$\\
			PAN~\cite{PAN/ZhengZY17aa} &$71.6$ & $83.9$ & $-$ & $90.6$ &$51.5$\\
			OIM~\cite{OIM/XiaoLWLW17} &$68.1$ & $-$ & $-$ & $-$ & $-$\\
			SVDNet~\cite{SVDNET/SunZDW17} &$76.7$ & $86.4$ & $89.9$ & $-$ & $56.8$\\
			AOS \cite{huang2018adversarially} &$79.2$ & $-$ & $-$ & $-$ &$62.1$\\
MLFN \cite{DBLP:journals/corr/abs-1803-09132} &$81.0$ & $-$ & $-$ & $-$ &$62.8$\\
HA-CNN \cite{li2018harmonious} &$80.5$ & $-$ & $-$ & $-$ &$63.8$ \\
PCB-RPP~\cite{sun2018beyond} &$83.3$ & $90.5$ & $92.5$ & $-$ & $69.2$\\
 \hline\hline
			\textbf{WBC-full-GoogLeNet} &  $76.2$ & $87.2$ & $91.1$ & $93.5$ &$56.9$\\
\textbf{WBC-full-Resnet} &  $81.7$ & $91.4$ & $93.1$ & $95.7$ &$69.1$\\
			\hline
		\end{tabular}
	\end{center}
\vspace{-2ex}
\end{table}

On the DukeMTMC-reID dataset, we compare our method with LOMO+XQDA, BoW, LSRO, ACRN~\cite{ACRN/SchumannS17}, PAN~\cite{PAN/ZhengZY17aa}, OIM~\cite{OIM/XiaoLWLW17}, SVDNet~\cite{SVDNET/SunZDW17}, Harmonious Attention Network (HA-CNN)~\cite{li2018harmonious}, AOS~\cite{huang2018adversarially}, Multi-Level Factorisation Net (MLFN)~\cite{DBLP:journals/corr/abs-1803-09132} and PCB-RPP~\cite{sun2018beyond}. and the comparison results are listed in Table~\ref{tab:duke_Dataset}.

As shown in Table~\ref{tab:duke_Dataset}, our approach based on Resnet backbones outperforms most of other state-of-the-art algorithms except a slightly lower performance than PCB-RPP~\cite{sun2018beyond}. But it has been proven that our framework with RPP in Table~\ref{table2} has better performance than PCB-RPP.

Only using GoogLeNet as backbone, our model's performance can beat other methods listed above SVDNet~\cite{SVDNET/SunZDW17} except a slightly lower rank-1 recognition rate compared to SVDNet~\cite{SVDNET/SunZDW17}. In ACRN~\cite{ACRN/SchumannS17}, the authors propose to separately learn a classifier to leverage the complementary information of attributes for better representation of the human appearance. The notable performance gain over ACRN~\cite{ACRN/SchumannS17} (3.6\% in rank 1 recognition rate and 4.9\% in mAP) clearly demonstrates that our approach can generate more representative appearance descriptors without the need for extra attribute annotations. Comparing Table~\ref{tab:duke_Dataset} with Table~\ref{tab:number_of_parts}, we can observe that our WBC model with one salient part ($L=1$ in Table~\ref{tab:number_of_parts} ) demonstrates superior performance than PAN~\cite{PAN/ZhengZY17aa}, reflecting the discriminative ability of the proposed \emph{weighted bilinear coding} model. The performance gain with bigger $L$ ($L=5$ in Table~\ref{tab:number_of_parts}) further validates the superiority of our salient part based representation over the global parameter aligned deep features utilized in PAN~\cite{PAN/ZhengZY17aa}. Our approach achieves slightly lower rank-1 recognition performance than SVDNet~\cite{SVDNET/SunZDW17} because SVDNet adopts ResNet-50 for feature extraction, which is more powerful than GoogLeNet in our approach.

\subsubsection{Results on CUHK03}

On the CUHK03 dataset, we follow the new evaluation protocol in~\cite{re-ranking/ZhongZCL17} to demonstrate the effectiveness of the proposed algorithm. We record the recognition performance on both settings of human-labeled and auto-detected bounding boxes and compare it with LOMO+XQDA, BOW+XQDA~\cite{market_dataset/ZhengSTWWT15}, IDE+DaF~\cite{bmvc_cuhk03}, PAN, DPFL~\cite{dpfl/ChenZG17}, Re-ranking, and SVDNet. Table~\ref{tab:cuhk03_Dataset} demonstrates the detailed comparison results. In this section, we product experiments based on GoogLeNet backbone.   

\renewcommand\arraystretch{1.2}
\begin{table}[!htbp]
	\centering
	\caption{Comparison results of top $r$ matching rate using CMC (\%) and mean average precision (mAP \%) on the CUHK03 dataset.
		\label{tab:cuhk03_Dataset}}
	\begin{tabular}{r|cc|cc}
		\hline
		\multirow{2}[4]{*}{Methods} & \multicolumn{2}{c|}{Labeled} & \multicolumn{2}{c}{Detected} \\
		\cline{2-5}          & r=1 & mAP   & r=1 & mAP \\
		\hline\hline
		LOMO+XQDA~\cite{lomo_xqda/LiaoHZL15}  & $14.8$  & $13.6$   & $12.8$   & $11.5$ \\
		BOW+XQDA~\cite{market_dataset/ZhengSTWWT15}  & $7.9$  &$7.3$   & $6.4$   & $6.4$ \\
		
		Re-ranking~\cite{re-ranking/ZhongZCL17} & $38.1$   & $40.3$   & $34.7$   & $37.4$ \\
		IDE+DaF~\cite{bmvc_cuhk03}& $27.5$   & $31.5$   & $26.4$   & $30.0$ \\
		PAN~\cite{PAN/ZhengZY17aa} &$36.9$   & $35.0$  & $36.3$   & $34.0$ \\
		DPFL~\cite{dpfl/ChenZG17}& $43.0$   & $40.5$   & $40.7$   & $37.0$ \\
		SVDNet~\cite{SVDNET/SunZDW17}& $40.9$  & $37.8$  & $41.5$   & $37.3$ \\
		\hline\hline	
		\textbf{WBC-full-GoogLeNet}  & $50.1$   & $47.7$  & $43.9$   & $42.1$ \\
		\hline
	\end{tabular}%
	\label{tab:addlabel}%
\end{table}%

As shown in Table~\ref{tab:cuhk03_Dataset}, under the 767/700 setting, our approach clearly outperforms the other state-of-the-arts. For example, on the labeled bounding-boxes, our approach outperforms the second best DPFL~\cite{dpfl/ChenZG17} by 7.1\% in rank-1 recognition rate and 7.2\% in mAP. Meanwhile on the detected bounding-boxes, we gain by 2.4\% in rank-1 recognition rate and 4.7\% in mAP. It is worth noting that PAN~\cite{PAN/ZhengZY17aa} is also an alignment net based deep feature learning model. Our approach reports to be superior than PAN~\cite{PAN/ZhengZY17aa} in both the labeled and detected settings. We attribute the obvious performance gain to two factors: (1) The richer higher-order information encoded in our \emph{weighted bilinear coding} model brings more representative ability to the deep features, and (2) The salient part network adopted in our model achieves more flexibility and better alignment performance than the globally parameterized spatial transformer network in PAN~\cite{PAN/ZhengZY17aa}. Another highly expected phenomenon on the CUHK03 dataset is that recognition performance on the manually-labeled images is indeed better than that of the auto-detected images, which reflects that more severe misalignment and noisy background clutters present in the auto-detected images need further consideration during the model design.

\begin{figure}[!t]
	\centering
	\includegraphics[width=\linewidth]{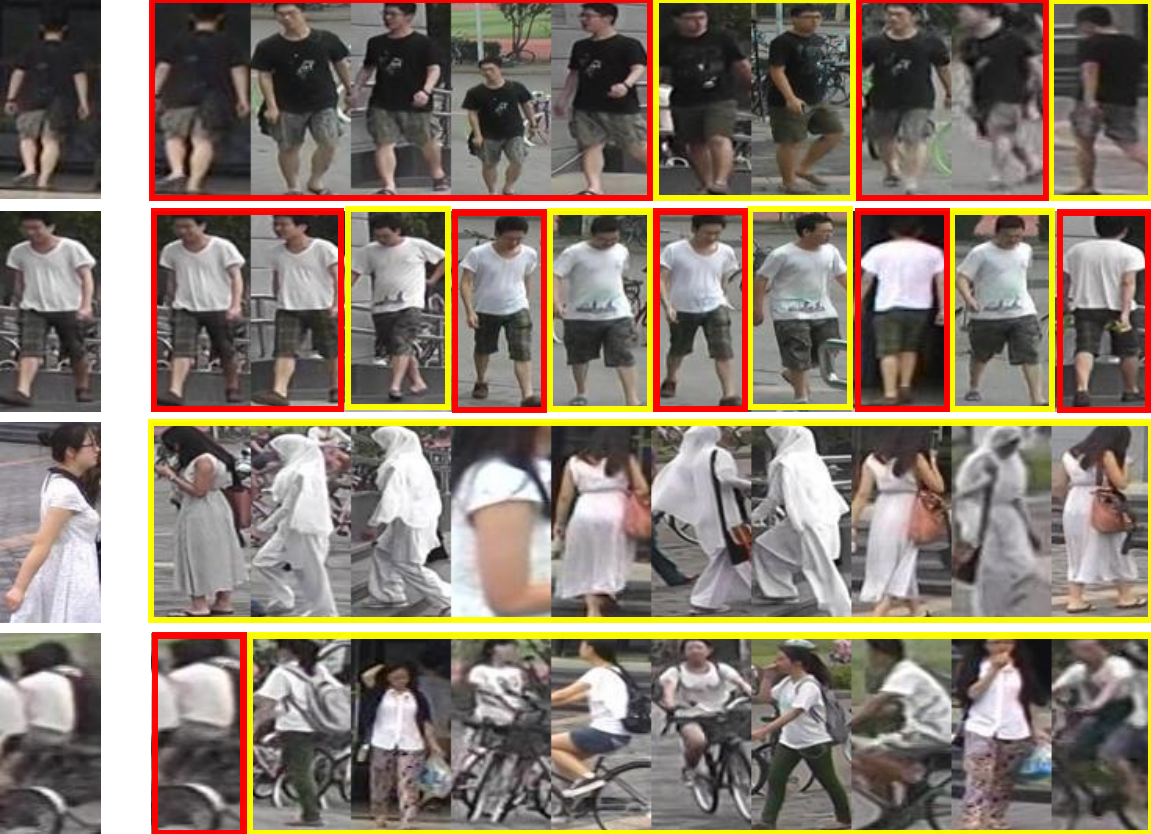}
	\caption{Instances of some typical failure cases. The first column are four probe images, and the following are ten most similar images generated by the proposed algorithm. Images in the red bounding boxes are true matches with the same identity, and images in the yellow bounding boxes are false positives with different identities. Generally, the proposed algorithm can rank visually similar images ahead of others. Best viewed in color.}
	\label{fig:failure_cases}
\end{figure}

\subsection{Typical Failure Cases Analysis}
\begin{figure*}[!t]
	\centering
	\includegraphics[width=0.99\linewidth]{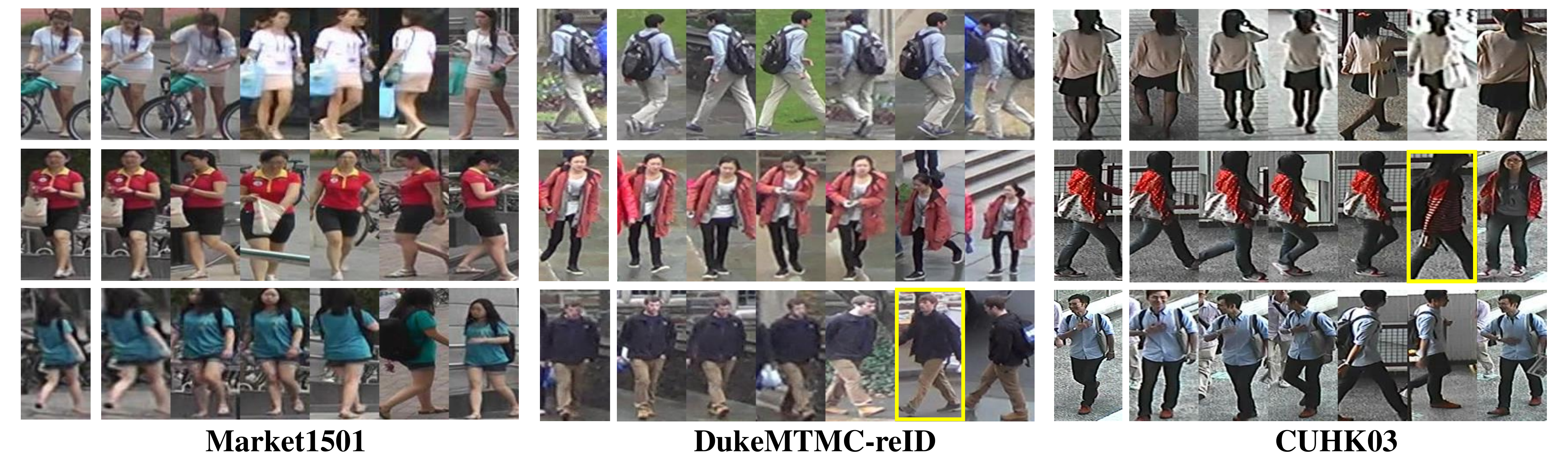}
	\captionsetup{font={footnotesize}}
	\vspace{-1ex}
	\caption{Example ranking results on the three benchmarks generated by our approach. On each dataset, the first columns are three probe images, followed by six top-ranked gallery images. Images marked by yellow bounding boxes are incorrect matches with different identities as the probes. Best viewed in color.}\label{fig:ranking_results}
	\vspace{-2ex}
\end{figure*}

In order to explore the limitations of the proposed algorithm, we present some typical failure cases on the Market-1501 dataset. As shown in Figure~\ref{fig:failure_cases}, there are mainly three factors that lead to undesirable matching results. The first two rows illustrate that when the negative instances are very similar with the probe image, false positives may be ranked ahead of true matches (e.g., the false positives in the first and second rows only slightly differ from the related probe images in the patterns on the T-shirts). In the third row, our algorithm does not find out the true matches within the top-10 rankings. This can be attributed to the partial detection result of the probe image, resulting in incomplete feature representation of the whole body. In the last row, blurring and serious background clutters misleads our algorithm into matching the probe with images in white upper clothes and on/beside a bicycle. In addition, we also choose some good examples ranking results in three datasets, which are demonstrated in Figure~\ref{fig:ranking_results}.

\section{Conclusions}
This paper proposes a novel \emph{weighted bilinear coding} (WBC) model to pursue more representative and discriminative aggregation for the intermediate convolutional features in CNN networks. In specific, channel-wise feature correlations are encoded to model higher-order feature interactions, improving the representative ability. Moreover, a weighting scheme is adopted to adaptively weigh local features to reflect local feature importance. Besides, to deal with spatial misalignment, a salient part net is introduced to automatically derive salient body parts. By integrating the WBC model and the salient part net, the final human appearance representation is both discriminative and resistant to spatial misalignment. Extensive experiments on three large-scale benchmarks demonstrate the effectiveness of the proposed approach. The flexibility and generalization of this framework also have been proven. 
\section*{Acknowledgments}
This work was supported by the National Natural Science Foundation of China (NSFC) (Grant No. 61671289, 61221001, 61528204, 61771303, 61521062 and 61571261), STCSM (18DZ2270700) and by US National Science Foundation (NSF) (Grant No.1618398 and 1350521). We gratefully acknowledge the support of NVIDIA Corporation with the donation of the Titan X Pascal GPU used for this research.


\end{document}